%% file: wikivqabench.tex
\documentclass{article}
\usepackage[eandd, preprint]{neurips_2026}
\usepackage[utf8]{inputenc}
\usepackage[T1]{fontenc}
\usepackage{hyperref}
\usepackage{url}
\usepackage{booktabs}
\usepackage{amsfonts}
\usepackage{nicefrac}
\usepackage{microtype}
\usepackage{xcolor}
\usepackage{graphicx}
\usepackage{listings}
\usepackage[caption=false]{subfig}
\usepackage{wrapfig}

\newcommand{\framework}{WikiVQABench}
\newcommand{\benchmarksize}{344}

\begin{document}

\title{\framework: A Knowledge-Grounded Visual Question Answering Benchmark from Wikipedia and Wikidata}

\author{
  Basel Shbita \\
  IBM Research \\
  San Jose, CA \\
  \texttt{basel@ibm.com} \\
  \And
  Pengyuan Li \\
  IBM Research \\
  San Jose, CA \\
  \texttt{pengyuan@ibm.com} \\
  \And
  Anna Lisa Gentile \\
  IBM Research \\
  San Jose, CA \\
  \texttt{annalisa.gentile@ibm.com}
}

\maketitle

\input{latex/abstract}

\sloppy

\input{latex/introduction}

\input{latex/related_work}

\input{latex/pipeline}

\input{latex/data}
\input{latex/evals}
\input{latex/conclusion}

\bibliographystyle{unsrtnat}
\bibliography{wikivqabench}

\appendix
\input{latex/appendix}


\end{document}

%% file: latex/abstract.tex
\begin{abstract}
Visual Question Answering (VQA) benchmarks have largely emphasized perception-based tasks that can be solved from visual content alone.
In contrast, many real-world scenarios require external knowledge that is not directly observable in the image to answer correctly.
We introduce \framework{}, a human-curated knowledge-grounded VQA benchmark constructed by systematically combining Wikipedia images, their associated article captions, and structured knowledge from Wikidata.
Our pipeline uses large language models (LLMs) to generate candidate multiple-choice image-question-answer sets.
All generated instances are subsequently reviewed and curated by human annotators to ensure factual correctness, visual-text consistency, and that each question requires external knowledge in addition to visual evidence for correct resolution.
\framework{} comprises a substantial collection of Wikipedia images with curated multiple-choice questions designed to benchmark knowledge-aware vision-language models (VLMs).
Evaluation of fifteen VLMs (256M-90B parameters) reveals a wide performance range (24.7\%-75.6\% accuracy), demonstrating that the benchmark effectively discriminates model capabilities on knowledge-intensive reasoning.
The dataset and benchmarking code are publicly available.
\end{abstract}

%% file: latex/introduction.tex
\section{Introduction}
\label{sec:intro}

Multimodal large language models (MLLMs) and vision-language models (VLMs) have recently shown strong capabilities across tasks involving vision and language, including image captioning, visual question answering (VQA), document understanding, and chart interpretation~\citep{liu2023visual,chen2024internvl,wang2024qwen2,rahmanzadehgervi2024vision,grattafiori2024llama,team2024gemini,dai2024nvlm,deitke2024molmo}.
Despite this progress, evaluating whether such models can reason beyond visual perception remains challenging, largely due to limitations in existing datasets and benchmarks.
The majority of VQA benchmarks currently used for evaluation are limited in scope.
Popular datasets focus on synthetic scenes, natural images with short factual queries, or narrow domains such as scientific diagrams and charts~\citep{li2025survey,tong2024eyes,duan2024vlmevalkit,zhang2024lmmsevalrealitycheckevaluation}.
These benchmarks often emphasize surface-level perception, such as object recognition or scene description, leaving open the broader challenge of knowledge-grounded visual cognition, where answering a question requires not only looking at an image but also understanding its real-world context, entities, and relationships.

Several datasets have attempted to incorporate external knowledge into VQA, including OK-VQA~\citep{marino2019ok}, A-OKVQA~\citep{schwenk2022okvqa}, and KVQA~\citep{shah2019kvqa}.
While these efforts represent important steps toward knowledge-aware evaluation, they exhibit notable limitations from a benchmarking perspective.
Many questions cover restricted sets of entities or domains or use open-ended answer formats that complicate standardized and reproducible evaluation.
More importantly, few existing benchmarks explicitly enforce that answering a question requires external knowledge beyond what can be inferred from the image itself, nor do they provide verifiable ground truth grounded in structured, machine-readable knowledge sources.

Real-world applications of VQA increasingly demand benchmarks that reflect these requirements.
Questions about historical artifacts, landmarks, public figures, or events often require recognizing entities in images and reasoning over their properties, relationships, and broader context using external knowledge.
Large, publicly available resources such as Wikipedia and Wikidata~\citep{vrandevcic2014wikidata} provide complementary visual, textual, and structured information that can support the construction of benchmarks where visual content serves as an anchor for entity-centric reasoning and answers can be traced back to explicit, verifiable facts.

To address this gap, we introduce \framework{}, a human-curated, knowledge-grounded visual question answering benchmark constructed by systematically combining Wikipedia images, associated article-image captions, and structured knowledge from Wikidata.
\framework{} is designed as a dataset and a benchmark to evaluate whether VLMs and MMLMs can integrate visual evidence with external, structured knowledge, rather than relying on visual perception alone.

\begin{wrapfigure}{r}{0.5\linewidth}
    \centering
    \includegraphics[width=0.85\linewidth]{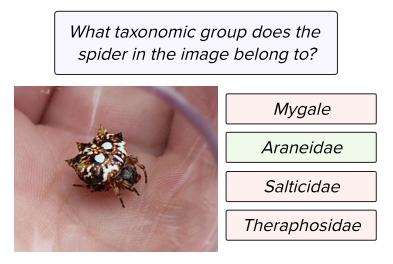}
    \caption{Example from \framework{} illustrating a knowledge-grounded multiple-choice VQA instance. 
    The image depicts a spider whose taxonomic classification cannot be determined from visual appearance alone. 
    Correctly answering the question requires external biological knowledge linking visual cues to entity-level taxonomy (e.g., family or genus), demonstrating the benchmark's emphasis on required knowledge beyond surface-level perception.}
    \label{fig:vqa_example}
\end{wrapfigure}

The dataset construction process combines automated generation with rigorous human curation.
Candidate multiple-choice image-question-answer sets are first generated using large language models (LLMs) conditioned on image captions and verbalized Wikidata triples.
These candidates are then reviewed and curated by human annotators to ensure factual correctness, visual-text consistency, and that each question cannot be answered from the image alone.
By grounding answers in structured knowledge and enforcing knowledge necessity through human curation, \framework{} provides verifiable ground truth suitable for standardized evaluation and comparative benchmarking.

\framework{} comprises \benchmarksize{} images paired with curated multiple-choice questions spanning a diverse set of entities, relations, and domains.
The benchmark is designed to support reproducible and scalable evaluation of knowledge-aware VLMs and MLLMs, enabling controlled analysis of entity-based and multi-hop ``understanding'' capabilities.
Figure~\ref{fig:vqa_example} illustrates a representative example from \framework{}, where answering the question requires taxonomic knowledge that cannot be inferred from visual appearance alone.
We open source \framework{}\footnote{Code available as part of \textit{VLMEvalKit} at: \url{https://github.com/open-compass/VLMEvalKit}} and make the benchmark dataset publicly accessible\footnote{Dataset available at: \url{https://huggingface.co/datasets/ibm-research/WikiVQABench}} in hopes of encouraging further work in developing more rigorous, fine-grained evaluation methodologies in this space.

\noindent
Our main contributions are as follows:
\begin{itemize}
  \item \textbf{A new knowledge-grounded VQA benchmark} leveraging Wikipedia and Wikidata that requires external, structured knowledge beyond visual perception to answer correctly.
  \item \textbf{A human-curated dataset construction methodology} that ensures factual correctness, verifiable ground truth, and enforced knowledge necessity.
  \item \textbf{Comprehensive evaluation across multiple state-of-the-art models}, demonstrating the benchmark's effectiveness in assessing knowledge-grounded visual reasoning capabilities.
  \item \textbf{An open and accessible benchmarking resource}, including dataset documentation, metadata, and tools.
\end{itemize}


%% file: latex/related_work.tex
\section{Related work}
\label{sec:related_work}

Visual Question Answering (VQA) has been studied extensively through a wide range of datasets and benchmarks.
Early benchmarks such as VQA-v2~\citep{goyal2017making} and GQA~\citep{hudson2019gqa} primarily focus on perception-based reasoning, emphasizing object recognition, spatial relations, and compositional visual understanding.
While these datasets have driven progress in visual reasoning, they are largely solvable using image content alone and do not require access to external knowledge.
To address this limitation, several benchmarks have introduced external knowledge into VQA.
Benchmarks like OK-VQA~\citep{marino2019ok}, A-OKVQA~\citep{schwenk2022okvqa}, and KVQA~\citep{shah2019kvqa} have explored entity-centric and knowledge-aware VQA, but exhibit limitations: knowledge requirements are often shallow or implicit, domain coverage is restricted, and open-ended answer formats complicate standardized and reproducible benchmarking.

More recent work has explored large-scale, knowledge-intensive VQA grounded in encyclopedic information.
Encyclopedic VQA~\citep{mensink2023encyclopedic} introduces a substantial collection of question-answer pairs (221k question-answer pairs and around 1M VQA samples) supported by evidence from a curated knowledge base derived from Wikipedia, demonstrating the importance of retrieval-augmented access to external knowledge.
EchoSight~\citep{yan2024echosight} proposes a retrieval-augmented generation framework that retrieves relevant Wikipedia articles based on visual input to support encyclopedic question answering.
While these approaches highlight the importance of external knowledge, they primarily focus on retrieval at inference time rather than on constructing benchmarks that explicitly enforce knowledge necessity and provide structured, verifiable ground truth.
Several datasets leverage Wikipedia as a source of visual and semantic information.
The Wikipedia-based Image Text (WIT) dataset~\citep{srinivasan2021wit} associates images with Wikipedia articles and captions, enabling large-scale image-text learning.
OVEN~\citep{hu2023open} introduces open-domain visual entity recognition by unifying image classification and QA datasets under a shared label space grounded in English Wikipedia, covering a broad range of entity types and granularities.
These datasets focus on entity recognition and visual linking, but they are not designed to evaluate whether models can reason over entity properties and relationships using external knowledge.

Structured knowledge graphs such as Wikidata~\citep{vrandevcic2014wikidata} provide machine-readable representations of factual knowledge and have been widely used in knowledge-grounded language generation and reasoning.
Wikidata, in particular, offers large-scale, multilingual coverage across diverse domains, making it well suited for dataset construction.
In the context of VQA, however, structured knowledge is often used implicitly or as auxiliary context, rather than as the basis for verifiable ground truth in benchmarking.

Recent advances in large language models have enabled scalable synthetic data generation for tasks such as instruction following and question synthesis~\citep{wang2025self}.
While automated generation enables scale, fully synthetic datasets often suffer from factual inaccuracies, visual inconsistencies, or ambiguous knowledge requirements.
As a result, there is growing recognition of the importance of combining automated generation with human curation to ensure data quality, correctness, and meaningful evaluation.
Our work builds on these lines of research by introducing a knowledge-grounded VQA benchmark that explicitly enforces the requirement of external knowledge beyond visual perception.
By systematically combining Wikipedia images, associated article captions, and structured knowledge from Wikidata, and by applying rigorous human curation, we provide a benchmark with verifiable ground truth suitable for standardized evaluation.

%% file: latex/pipeline.tex
\section{Dataset Construction and Pipeline}
\label{sec:pipeline}

\begin{figure*}[t]
    \centering
    \includegraphics[width=\linewidth]{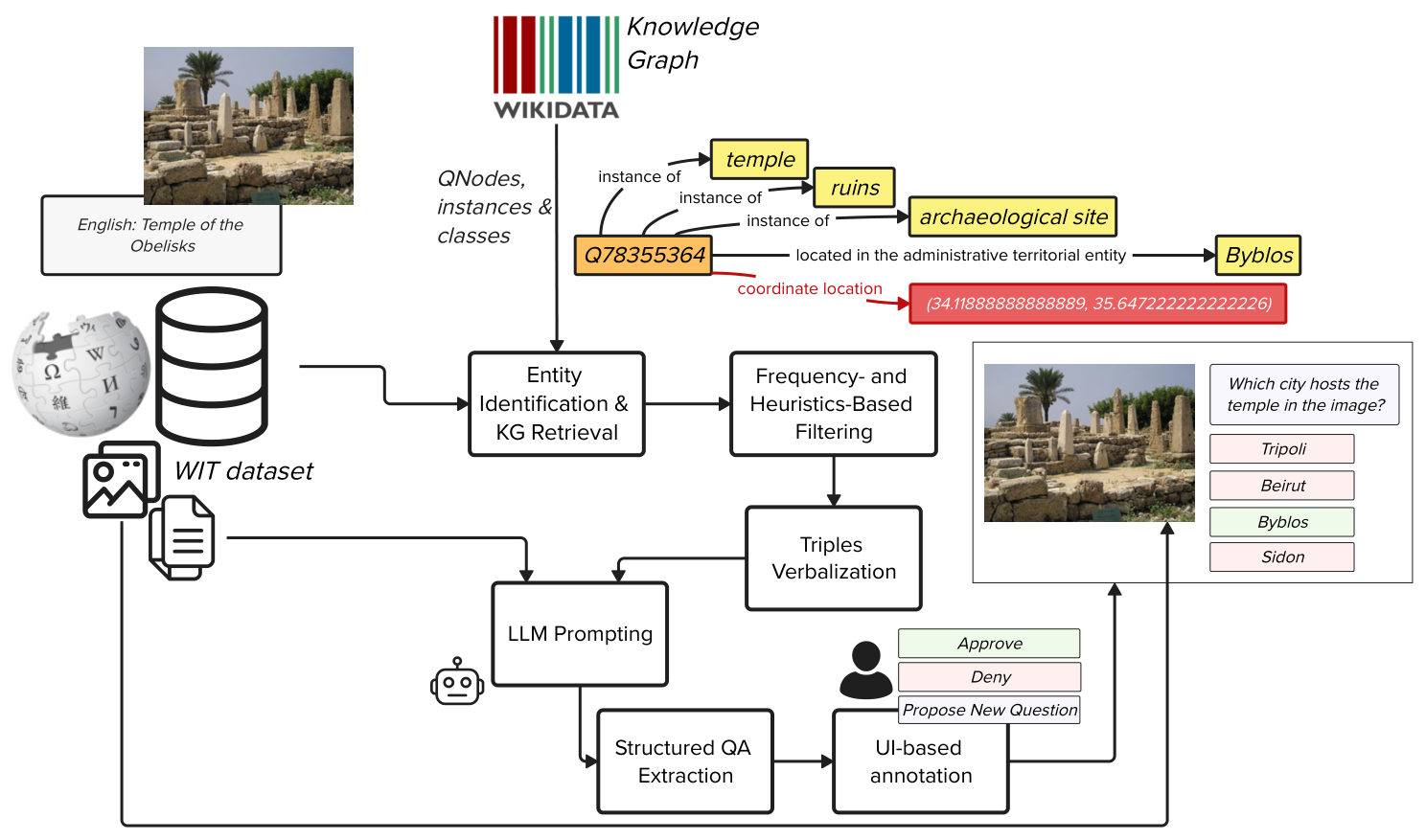}
    \caption{Overview of the \framework{} dataset construction pipeline.
    Left: image and caption input from the WIT dataset.
    Top (left to right): retrieval of Wikidata entities and associated triples (orange: matched entity; yellow: retained factual relations; red: filtered low-salience or metadata relations).
    Bottom: generation of candidate multiple-choice visual question-answer instances using verbalized triples and image captions, followed by human review and curation to ensure factual correctness and enforced knowledge necessity.}
    \label{fig:data_pipeline}
\end{figure*}

Our method generates knowledge-grounded visual question-answer instances by aligning Wikipedia images with structured Wikidata knowledge. We combine automated data generation with human curation to ensure factual correctness, visual-text consistency, and enforced knowledge necessity. Figure~\ref{fig:data_pipeline} illustrates the overall pipeline used to construct \framework{}.

We build on the Wikipedia-based Image Text (WIT) dataset~\citep{srinivasan2021wit}, a large collection of over 37 million image-text associations extracted from Wikipedia articles. Each instance includes an image, its associated caption, and metadata linking the image to a specific Wikipedia article. These article links provide the foundation for entity-centric knowledge retrieval and enable scalable alignment between visual content and structured knowledge.

The construction process begins with an image-caption pair from WIT (left side of Figure~\ref{fig:data_pipeline}). Using Wikipedia page metadata, we identify entities mentioned in the associated article and resolve them to Wikidata identifiers (QNodes), which serve as anchors for structured knowledge retrieval (Section~\ref{sec:entity_linking}). For each resolved entity, we retrieve factual relations and descriptive attributes from Wikidata. We explicitly discard relations involving protected attributes, such as age, gender, sexual orientation, race (including color, nationality, ethnic or national origins), religion, beliefs, and religious practices.\footnote{\url{https://www.legislation.gov.uk/ukpga/2010/15}}
These triples may connect entities to other entities, classes, or literal values, enabling multi-hop reasoning chains.

Not all retrieved relations are equally suitable for visual question answering. To reduce noise and improve data quality, we apply frequency- and heuristic-based filtering to remove overly generic, metadata-oriented, or weakly informative predicates (Section~\ref{sec:filter_triples}). This step prioritizes semantically meaningful relations that are more likely to support visually grounded questions while improving consistency and reducing annotation burden.
The remaining triples are verbalized as natural language statements and combined with the original image caption to form a structured textual context. This context is provided to an LLM-based generator, which produces candidate multiple-choice visual question-answer instances grounded in both the image caption and the retrieved knowledge (Section~\ref{sec:qa_gen}). These instances are then normalized into a structured, machine-readable representation to support systematic review and downstream processing (Section~\ref{sec:reformat}).

All automatically generated instances undergo human review through a UI-based annotation process. Annotators verify factual correctness, visual-text consistency, and ensure that the correct answer requires external knowledge rather than relying solely on surface-level visual cues (Section~\ref{sec:human_curation}). Annotators may approve, reject, or revise questions to improve clarity and quality.
The final output of this pipeline is \framework{}, a curated benchmark of multiple-choice visual question-answer pairs grounded in structured knowledge. By combining Wikipedia images, Wikidata triples, automated generation, and human verification, the dataset enables reproducible, scalable evaluation of knowledge-aware vision-language models.


\subsection{Entity Identification and KG Retrieval}
\label{sec:entity_linking}

For each image in the WIT dataset, we first identify the source Wikipedia article using a rule-based URL extraction procedure.
From the article URL, we retrieve the corresponding Wikidata item (QNode) via the Wikipedia API\footnote{\url{https://www.wikidata.org/w/api.php}}, which uniquely identifies the primary entity associated with the image.
Once the QNode is resolved, we retrieve all subject-predicate-object triples directly connected to that entity by issuing SPARQL~\citep{world2013sparql} queries to the Wikidata endpoint\footnote{\url{https://query.wikidata.org/}}.
At this stage, we collect an unrestricted set of outgoing triples without applying filtering or heuristics.
This raw triple set captures a broad range of factual assertions, including class membership, taxonomic relations, geographic information, and descriptive attributes.
The resulting triples form the input to the filtering stage described next.


\subsection{Filtering and Verbalizing Wikidata Triples}
\label{sec:filter_triples}

Not all retrieved Wikidata relations are suitable for knowledge-grounded visual question answering.
To guide the design of our filtering strategy, we conducted a corpus-level analysis of predicate and object frequencies over a random sample of approximately 400k Wikidata entities aligned with WIT images.
This analysis revealed that the majority of Wikidata facts associated with Wikipedia images concentrate around a small number of high-support, human-interpretable semantic relations.

\begin{wraptable}{r}{0.45\linewidth}
\vspace{-0.2in}
    \small
    \centering
    \caption{Most notable frequent predicates and objects observed in our 400k-entity sample. These statistics inform our filtering strategy by highlighting relations that are common, human-interpretable, and often visually grounded.}
    \label{tab:top_predicates_objects}
    \begin{tabular}{llr}
    \toprule
    \textbf{Qnode} & \textbf{Label} & \textbf{Count} \\
    \midrule
    P31   & instance of & 846,144 \\
    P625  & coordinate location &  320,776 \\
    P106  & occupation &  165,531 \\
    P279  & subclass of & 123,876 \\
    P166  & award received &   55,738 \\
    \midrule
    Q5        & human & 267,990 \\
    Q16521    & taxon &  62,940 \\
    Q4830453  & business &  14,734 \\
    Q33999    & actor &  11,259 \\
    Q515      & city  &  8,850 \\
    \bottomrule
    \end{tabular}
\vspace{-0.1in}
\end{wraptable}

The distributions in the Table~\ref{tab:top_predicates_objects} show that a large fraction of relevant facts fall into a few dominant semantic categories, including biographical (e.g., \texttt{human (Q5)}, \texttt{occupation (P106)}), geographic (e.g., \texttt{city (Q515)}, \texttt{coordinate location (P625)}), and taxonomic relations (e.g., \texttt{taxon (Q16521)}, \texttt{subclass of (P279)}).
These relations are frequently interpretable by humans and often align well with visually grounded content, making them suitable for constructing knowledge-required VQA instances.

Guided by this analysis, we apply a unified filtering strategy that combines predicate frequency thresholds with heuristic removal of low-value metadata properties.
First, we retain only predicates with sufficient support in the WIT-aligned corpus, discarding long-tail relations that occur fewer than 10 times and rarely contribute meaningful visual grounding.
Second, we explicitly remove administrative or identifier-oriented predicates whose labels match patterns such as \texttt{ID}, \texttt{identifier}, \texttt{number}, \texttt{code}, \texttt{username}, \texttt{URL}, or \texttt{website}.
Although such properties are common in Wikidata, they provide little semantic value for visual reasoning and are not visually grounded.

All the remaining triples after this filtering process are verbalized into natural language statements. The verbalization process consists of taking the triple ($t$) <subject ($s$), predicate ($p$), object ($o$)> and generating a sentence that expresses $t$. The verbalization involves resolving all $s$, $p$, and $o$ using their literal labels, or resolving nested references that can occur in the objects.
For example, in the Wikidata triple:

\begin{lstlisting}
s: Q217099
p: P2043
o: wikidata.quantity.Quantity(1300.0, None, 
   None, <wikidata.entity.Entity Q828224>)
\end{lstlisting}
we can easily resolve $s$ and $p$ by fetching their labels - \texttt{Q217099} is ``\texttt{Karakoram Highway}'' and \texttt{P2043} is ``\texttt{length}'', while to resolve $o$ we need to first resolve the Wikidata entity \texttt{Q828224} to understand that the expressed quantity is in ``\texttt{kilometre}''.
The verbalization for this specific example would be:

\begin{lstlisting}
Detected QNode: Karakoram Highway( Q217099 )
Filtered triples ( verbalized ):
- length: "1300 kilometres".
- ... <all other extracted triples> 
\end{lstlisting}

These verbalized triples form a human-readable representation of structured knowledge that can be reliably consumed by downstream generation components.


\subsection{LLM-Based Question-Answer Generation}
\label{sec:qa_gen}


Following triple verbalization, we generate candidate multiple-choice visual question--answer (VQA) instances using \textit{Granite-3.3-8B-Instruct}~\citep{granite2024granite}, a compact, open-source LLM designed for efficiency.
For each image, we construct a structured textual context by combining the image caption with the set of verbalized Wikidata triples retained after filtering.
We use a constrained prompt that instructs the LLM to generate single-sentence questions paired with one correct answer and three incorrect distractors.
The prompt enforces that questions are directly related to the image content while remaining consistent with the provided factual context.
Importantly, the prompt discourages explicit references to the source caption or knowledge triples, encouraging the model to synthesize coherent knowledge-informed questions rather than restating inputs.
The base prompt used to guide this generation is shown in Listing~\ref{lst:gen_prompt}.

\begin{lstlisting}[
    label=lst:gen_prompt,
    basicstyle=\ttfamily\scriptsize,
    frame=single,
    caption={prompt used to guide the LLM in generating knowledge-grounded QA pairs from image captions and verbalized triples.},
    captionpos=b,
    numbers=none,
    breaklines=true,
    breakindent=0pt,
    showstringspaces=false
]
Generate a set of single sentence questions with their multiple choice single word answers (one is correct, three are incorrect) about an image, given the image caption and a set factual elements in the form of verbalized knowledge triples that follow. The questions should be directly related to the image. Do not refer to the image caption or the verbalized knowledge directly. Format output as follows:
<##Question##> Your question here
<##Correct_Answer##> correct_answer
<##Wrong_Answer##> wrong_answer_1
<##Wrong_Answer##> wrong_answer_2
<##Wrong_Answer##> wrong_answer_3
\end{lstlisting}

Since all verbalized facts are explicitly grounded in Wikidata entities and relations, the generated questions and answers remain traceable to structured knowledge.
This allows the LLM to combine multiple facts, for example by identifying the depicted entity and reasoning over its properties or relationships, while still producing candidates that can be systematically reviewed and validated.


\subsection{Structured Representation for Review}
\label{sec:reformat}

To support human review and downstream dataset assembly, we normalize the raw LLM outputs into a consistent, structured representation.
This step prepares each candidate VQA instance for inspection in a custom User-Interface (UI) and ensures uniform formatting across the dataset.

We first clean the generated text by removing auxiliary markers such as conversation tags, Markdown fences, and formatting artifacts.
Role markers from different prompting styles (such as ``\texttt{AI:}'', ``\texttt{assistant:}'', or special tokens like ``\texttt{<|user|>}'') are normalized using a set of deterministic regular-expression rules.
Valid outputs are then mapped into a JSON-style schema that explicitly separates the question, correct answer, and incorrect answers.
Listing~\ref{lst:qa_restruct_example} shows a complete example. 

\begin{lstlisting}[
    label=lst:qa_restruct_example,
    basicstyle=\ttfamily\scriptsize,
    frame=single,
    caption={Example of a normalized, structured VQA instance produced by the pipeline.
The listing shows the detected Wikidata entity, the filtered and verbalized triples used as factual grounding, the original image caption, and the final multiple-choice question-answer instance after normalization.},
    captionpos=b,
    numbers=none,
    breaklines=true,
    breakindent=0pt,
    showstringspaces=false
]
Detected QNode:  "Obelisk Temple" (Q78355364)
Image caption:   "English: Temple of the Obelisks"
Filtered triples (verbalized):
  - "instance of": "temple".
  - "instance of": "ruins".
  - "instance of": "archaeological site".
  - "located in the administrative territorial entity": "Byblos".
Extracted QA:
{ "question": "Which city hosts the temple in the image?",
  "correct_answer": "Byblos",
  "wrong_answers": ["Tripoli", "Beirut", "Sidon"] }
\end{lstlisting}



\subsection{Human Curation and Quality Control}
\label{sec:human_curation}

All automatically generated VQA instances undergo human review through a UI-based annotation process.
Annotators are presented with the image, the generated question, and the associated answer options, with the correct answer indicated by the generation pipeline.
For each instance, annotators assess factual correctness, visual-text consistency, and whether answering the question genuinely requires knowledge beyond surface-level visual cues.

Annotators may approve instances that meet all criteria, reject instances that contain factual errors or are answerable from the image alone, or propose revised questions that better enforce knowledge necessity.
Out of 2,369 total instances reviewed, \benchmarksize{} (14.5\%) were accepted or accepted with suggested revisions, while 2,025 (85.5\%) were rejected due to factual inaccuracies, insufficient knowledge requirements, or visual-text inconsistencies.
This stringent quality control reflects the high bar we set for knowledge necessity and correctness.
Further details and a screenshot of the UI we have developed for this task is included in Section~\ref{sec:appndx_human_curation}.
This workflow streamlines annotation, reduces cognitive load, and enables consistent quality control across annotators by making grounding signals and decision options explicit.
Having a human-in-the-loop serves as a final quality control stage, ensuring that the resulting benchmark reflects meaningful, knowledge-grounded visual reasoning rather than superficial perception or spurious correlations.

%% file: latex/data.tex
\section{Dataset Characteristics}
\label{sec:data}


\begin{wraptable}{r}{0.4\linewidth}
\vspace{-0.25in}
    \small
    \centering
    \caption{Question Type Distribution}
    \begin{tabular}{lrr}
    \toprule
    Question Type & Count & Percentage \\
    \midrule
    Which & 160 & 46.5\% \\
    What & 133 & 38.7\% \\
    Who & 13 & 3.8\% \\
    When & 9 & 2.6\% \\
    Where & 7 & 2.0\% \\
    Other & 22 & 6.4\% \\
    \midrule
    Total & 344 & 100.0\% \\
    \bottomrule
    \end{tabular}
    \label{tab:question_types}
\vspace{-0.1in}
\end{wraptable}

\paragraph{Question Type.}

We classify questions into types based on question words: \emph{Which}, \emph{What}, \emph{Who}, \emph{When}, \emph{Where}, and \emph{Other}.
Table~\ref{tab:question_types} shows the distribution of question types across the benchmark.
The diversity of question types ensures that the benchmark tests multiple capabilities, mirroring the encyclopedic nature of knowledge grounding and emphasizes entity- and attribute-centric reasoning over open-ended description.
The \emph{Other} category captures a mixed set of structured but non-canonical interrogative forms that do not lexically begin with a ``standard'' question word.
The classification is purely lexical and based on the first word of the question, rather than on the presence of interrogative terms elsewhere in the sentence.
As a result, many questions in the \emph{Other} category still contain interrogative words such as \emph{which} or \emph{what} later in the sentence (e.g., ``\texttt{At which institution was this photograph taken?}'').


\paragraph{Answer Type.}

We categorize answers into three primary types: \emph{Descriptive}, \emph{Numeric}, and \emph{Alphanumeric}.
In total, there are 1,087 \emph{Descriptive} answers; these include short textual labels and named entities such as names of animal species (e.g., ``\texttt{Sixgill Hagfish}'', ``\texttt{Little Skate}'', ``\texttt{Leatherback Turtle}'') or object attributes.
There are 179 \emph{Numeric} answers, which primarily correspond to years or counts (e.g., ``\texttt{1987}'', ``\texttt{1990}'', ``\texttt{1980}''), supporting temporal and quantitative reasoning.
Finally, there are 110 \emph{Alphanumeric} answers, which consist of structured identifiers derived from curated knowledge sources, such as MeSH tree codes (e.g., ``\texttt{D12.776.157.530.450.437}''), which require precise knowledge rather than surface-level inference.



\begin{wraptable}{r}{0.5\linewidth}
\vspace{-0.25in}
    \small
    \centering
    \caption{Semantic Category Distribution}
    \begin{tabular}{lrr}
    \toprule
    Category & Count & Percentage \\
    \midrule
    Object/Thing & 90 & 26.1\% \\
    Location & 89 & 25.9\% \\
    Knowledge Identifier & 64 & 18.6\% \\
    Date/Time & 50 & 14.5\% \\
    Person & 48 & 14.0\% \\
    Other & 3 & 0.9\% \\
    \midrule
    Total & 344 & 100.0\% \\
    \bottomrule
    \end{tabular}
    \label{tab:semantic_categories}
\vspace{-0.1in}
\end{wraptable}

\paragraph{Semantic Content.}

We classify questions into semantic categories based on the type of knowledge required: \emph{Location} (geographical/spatial), \emph{Object/Thing}, \emph{Person}, \emph{Date/Time}, \emph{Knowledge Identifier}, and \emph{Other}.
This categorization reveals the diversity in question subjects.
Table~\ref{tab:semantic_categories} demonstrates that \textit{Object}-based questions constitute the largest category (26.1\%), followed by \textit{Location} identification (25.9\%), and \textit{Knowledge Identifier} questions (18.6\%).
These are questions requiring the retrieval of specific identifiers, taxonomic classes, or controlled-vocabulary entries.

Each semantic category corresponds to a distinct type of knowledge requirement.
\textit{Object/Thing} questions target properties or attributes of depicted items (e.g., ``\texttt{What is the material of the coin depicted in the image?}'').
\textit{Location} questions involve geographic or spatial grounding (e.g., ``\texttt{Which country is known as the place of origin for this sport?}'').
The \textit{Knowledge Identifier} category captures questions that require retrieving structured identifiers or taxonomic assignments (e.g., MeSH tree codes or biological genus).
\textit{Person}-centered questions query biographical facts such as occupation or associated activity. 
\textit{Date/Time} questions require temporal knowledge (e.g., birth year of a well-known entity depicted in the image).
The remaining category, named \textit{Other}, encompasses other question types that require other skills, such as quantitative reasoning or comparisons involving multiple entities.

%% file: latex/evals.tex
\section{Evaluations and Discussion}
\label{sec:evals}

Our evaluations with \framework{} are aimed at demonstrating its utility for probing vision-language models on knowledge-grounded visual question answering tasks.
We evaluate a diverse set of VLMs across different scales and architectures, establishing baseline results that reveal both the capabilities and limitations of current VLMs on knowledge-intensive reasoning.
This systematic evaluation allows us to identify capability gaps and scaling effects that are often overlooked in standard VQA benchmarks.


\paragraph{Experiments.}

We evaluate fifteen vision-language models representing different scales and architectures from different model families and variants from \textit{Granite-Vision}~\citep{team2025granite}, \textit{Qwen}~\citep{xu2025qwen2}, \textit{SmolVLM}~\citep{marafioti2025smolvlm}, \textit{Llama}~\citep{grattafiori2024llama}, \textit{Claude}~\citep{anthropic_claude}, and 
\textit{InternVL}~\citep{chen2024expanding}.
This selection spans from lightweight models (256M parameters) to large models (90B parameters), providing a balanced perspective on both architectural differences and scaling effects across VLM families.
All models are evaluated on the same \benchmarksize{} questions with accuracy as the primary metric.


\begin{wraptable}{r}{0.5\linewidth}
\vspace{-0.2in}
    \small
    \centering
    \caption{Model performance on \framework{}, ranked from highest to lowest score.}
    \begin{tabular}{lr}
    \toprule
    Model & Accuracy \\
    \midrule
InternVL3-78B & 75.6\% \\
Claude-Opus-4-6 & 70.3\% \\
Claude-Sonnet-4-6 & 66.3\% \\
Llama-3.2-90B-Vision-Instruct & 65.7\% \\
Qwen3-VL-32B-Instruct & 64.0\% \\
Qwen3-VL-8B-Instruct & 63.1\% \\
Qwen3-VL-4B-Instruct & 60.2\% \\
Qwen3-VL-2B-Instruct & 56.4\% \\
Granite-Vision-3.3-2B & 54.7\% \\
SmolVLM2 & 54.1\% \\
SmolVLM & 46.5\% \\
SmolVLM2-500M & 36.6\% \\
SmolVLM2-256M & 32.3\% \\
SmolVLM-500M & 29.4\% \\
SmolVLM-256M & 24.7\% \\
    \bottomrule
    \end{tabular}
    \label{tab:model_performance}
\end{wraptable}

\paragraph{Model Performance Analysis.}

Table~\ref{tab:model_performance} presents the overall accuracy for each model on \framework{}.
The results reveal significant performance variation across models, with the top-performing model (\textit{InternVL3-78B}) achieving 75.6\% accuracy, while the smallest variant (\textit{SmolVLM-256M}) achieves only 24.7\%.
This performance gap highlights both the role of model scale in knowledge-intensive reasoning and the challenge posed by our benchmark.

The 51 percentage-point gap between the strongest and weakest performers underscores the substantial challenge posed by knowledge-grounded reasoning on \framework{}.
Notably, models in the 8B-90B range cluster between 63\% and 66\% accuracy (\textit{Claude-Sonnet-4-6}, \textit{Llama-3.2-90B}, \textit{Qwen3-VL-32B/8B}), suggesting a performance plateau where additional scale yields diminishing returns without architectural advances in knowledge integration.

The overall accuracy range (24.7\%-75.6\%) indicates that \framework{} effectively discriminates between models while remaining genuinely challenging even for larger variants.
Critically, the smallest models barely exceed random chance (25\% for 4-way multiple choice), with the 256M variants achieving just 24.7\% and 32.3\% accuracy.
This indicates that scaling alone is insufficient: only the largest model exceeds 75\% accuracy, reflecting the demanding nature of integrating visual recognition with external structured knowledge, a capability not yet mature in current vision-language models.

\begin{wrapfigure}{r}{0.6\linewidth}
    \centering
    \includegraphics[width=\linewidth]{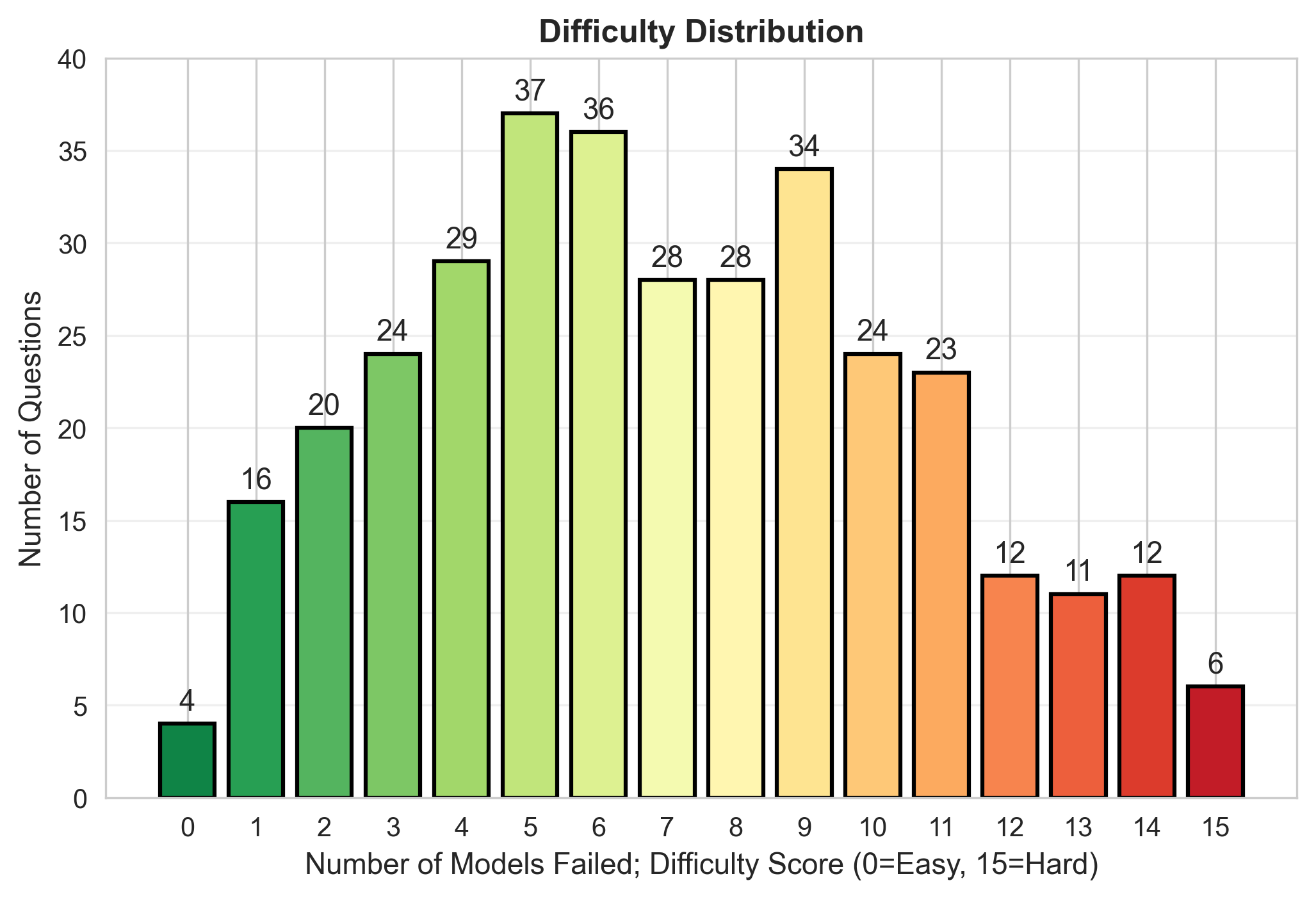}
    \caption{Question difficulty distribution. Each bar shows the number of questions for which exactly $k$ out of 15 models failed (mean = 7.0).}
    \label{fig:difficulty}
\vspace{-0.15in}
\end{wrapfigure}

\paragraph{Question Difficulty Analysis.}
We assess question difficulty by analyzing the collective performance of all fifteen models across the benchmark.
We compute a difficulty score for each question as the number of models that fail to answer it correctly, on a scale from 0 (all models solve) to 15 (no model solves).
The distribution of question difficulty (Figure~\ref{fig:difficulty}) demonstrates that our benchmark effectively challenges VLMs across multiple tiers.
The mean difficulty score of 7.0 indicates that \framework{} maintains an appropriate balance between accessibility and challenge.
With 4 questions solved by all models (easy tier) and 6 solved by none (hard tier), the benchmark effectively spans the full spectrum of model capabilities, enabling fine-grained discrimination between strong and weak performers.
The distribution is roughly unimodal, centered around medium difficulty, which is ideal for benchmarking.

%% file: latex/conclusion.tex
\section{Conclusion}
\label{sec:conclusions}

We introduced \framework{}, a benchmark for evaluating vision-language model capabilities on knowledge-grounded visual question answering tasks.
Our benchmark comprises \benchmarksize{} questions derived from Wikipedia and Wikidata, featuring diverse semantic categories (e.g., \textit{Location}, \textit{Object/Thing}, \textit{Person}, \textit{Date/Time}) and a carefully curated distractor set that requires genuine visual reasoning.
Evaluation of fifteen VLMs spanning different scales and architectures reveals significant performance gaps: the top model (\textit{InternVL3-78B}) achieves 75.6\% accuracy, while the smallest variant (\textit{SmolVLM-256M}) achieves only 24.7\%.
The question difficulty analysis shows a well-balanced distribution with a mean difficulty score of 7.0 (out of 15), ensuring effective discrimination between models while remaining appropriately challenging.
These findings underscore the need for specialized benchmarks like \framework{} to expose fine-grained weaknesses in knowledge-intensive reasoning and guide future VLM model development toward stronger semantic understanding.


\paragraph{Future Work.}

Looking ahead, several directions stand out.
First, scaling the benchmark to include more questions would broaden its applicability and provide more granular analysis across semantic categories and difficulty tiers.
Second, exploring specialized architectures designed for knowledge-intensive reasoning could reveal whether the current performance gaps are fundamental limitations or architecture-specific issues.
Finally, incorporating open-ended question answering and multi-turn dialogue would enable richer assessment of reasoning capabilities beyond multiple-choice selection.
Together, these extensions would strengthen the benchmark for systematically probing the limits of VLMs in knowledge-grounded visual understanding.


\paragraph{Limitations.}

Our proposed benchmark has several limitations worth noting.
First, while the \benchmarksize{} curated questions represent the result of rigorous human filtering (14.5\% acceptance rate from 2,369 generated instances), this focused size prioritizes quality and enforces strong knowledge necessity constraints over breadth.
The benchmark's focus on Wikipedia and Wikidata entities may limit generalizability to domains less well-represented in these knowledge bases, though these sources provide comprehensive coverage of notable entities across most domains.
Second, although our evaluation spans diverse model scales (256M to 90B parameters), the benchmark's \benchmarksize{} questions may limit statistical power for fine-grained per-category comparisons.
Third, the multiple-choice format, while enabling reproducible and standardized evaluation, may not capture all nuances of open-ended reasoning.
Finally, LLM-assisted generation may introduce subtle biases, though human curation mitigates this risk through explicit quality control.
These factors suggest natural extensions for future work.

%% file: latex/appendix.tex
\section{Human Curation Quality Control}
\label{sec:appndx_human_curation}

\begin{figure}[ht]
    \centering
    \includegraphics[width=\linewidth]{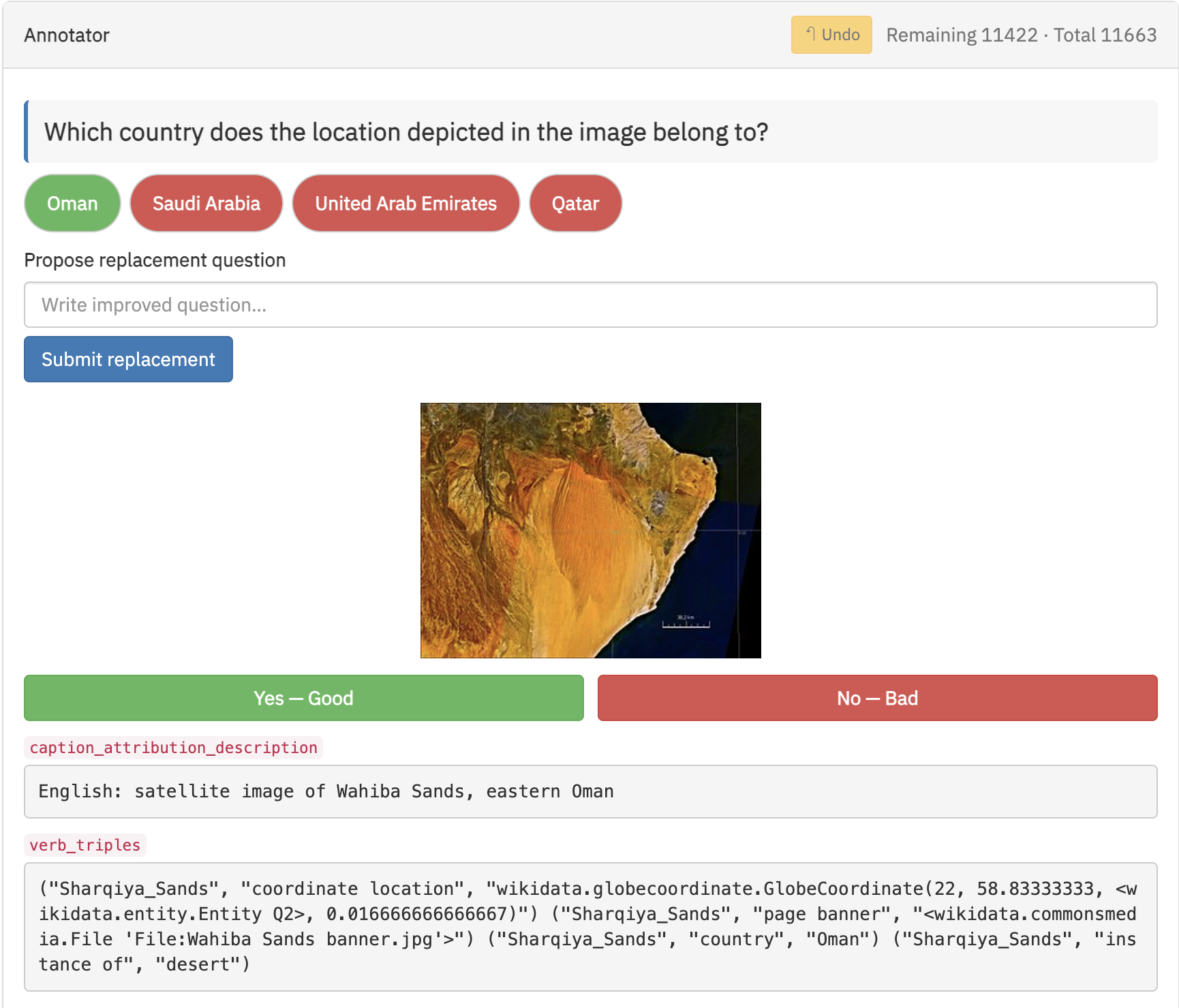}
    \caption{Screenshot of the UI used for human curation and quality control.
    Annotators are shown the image, the generated multiple-choice question, and the answer options.
    The correct answer is highlighted in green, while incorrect distractors are shown in red, making grounding and correctness explicit at a glance.
    Annotators can approve valid instances, reject incorrect or insufficiently grounded ones, or propose revised questions to better enforce knowledge necessity and visual-text consistency.}
    \label{fig:ui_screenshot}
\end{figure}

As mentioned in Section~\ref{sec:human_curation} the final set of samples included in the \framework{} was human curated for quality control where we verified the knowledge necessity and correctness for each sample.
Figure~\ref{fig:ui_screenshot} shows a snippet of the UI developed to support this annotation process.
The interface presents annotators with the image, the generated question, and the full set of answer options, with the correct answer explicitly marked.

